\newcommand{\q}{\boldsymbol{q}}
\newcommand{\qd}{\dot{\boldsymbol{q}}}
\newcommand{\qm}{\boldsymbol{q}_{{m}}}
\newcommand{\qb}{\boldsymbol{q}_{{b}}}

\newcommand{\qbd}{\dot{\boldsymbol{q}}_{{b}}}

\documentclass[a4paper,fleqn]{cas-dc}
\usepackage[numbers]{natbib}

\usepackage{graphicx}               
\usepackage{subcaption}             

\begin{document}
\let\WriteBookmarks\relax
\def\floatpagepagefraction{1}
\def\textpagefraction{.001}

\shorttitle{Safe Expeditious Whole-Body Control of Mobile Manipulators for Collision Avoidance}
\shortauthors{Bingjie Chen et~al.}

\title [mode = title]{Safe Expeditious Whole-Body Control of Mobile Manipulators for Collision Avoidance}                      



\author[1]{Bingjie Chen}[orcid=0009-0002-7926-2507]
\author[1]{Yancong Wei}
\author[3]{Rihao Liu}
\author[1]{Chenxi Han}
\author[1]{Houde Liu}
\cormark[1]
\author[2]{Chongkun Xia}
\author[3]{Liang Han}
\author[1]{Bin Liang}

\affiliation[1]{organization={Shenzhen International Graduate School, Tsinghua University},
                city={Shenzhen},
                postcode={518000},
                country={China}}
                
\affiliation[2]{organization={School of Advanced Manufacturing, Sun Yat-Sen University, Sun Yat-Sen University},
                city={Shenzhen},
                postcode={518000},
                country={China}}

\affiliation[3]{organization={School of Electrical and Automation Engineering, Hefei University of Technology},
                city={Hefei},
                postcode={230009},
                country={China}}

\nonumnote{* Corresponding author.}
\nonumnote{ E-mail address: cbj23@mails.tsinghua.edu.cn (B. Chen).}

\begin{abstract}
Whole-body reactive obstacle avoidance for mobile manipulators (MM) remains an open research problem. Control Barrier Functions (CBF), combined with Quadratic Programming (QP), have become a popular approach for reactive control with safety guarantees. However, traditional CBF methods often face issues such as pseudo-equilibrium problems (PEP) and are ineffective in handling dynamic obstacles. To overcome these challenges, we introduce the Adaptive Cyclic Inequality (ACI) method. ACI takes into account both the obstacle’s velocity and the robot’s nominal control to define a directional safety constraint. When added to the CBF-QP, ACI helps avoid PEP and enables reliable collision avoidance in dynamic environments. We validate our approach on a MM that includes a low-dimensional mobile base and a high-dimensional manipulator, demonstrating the generality of the framework. In addition, we integrate a simple yet effective method for avoiding self-collisions, allowing the robot enabling comprehensive whole-body collision-free operation. Extensive benchmark comparisons and experiments demonstrate that our method performs well in unknown and dynamic scenarios, including difficult tasks like avoiding sticks swung by humans and rapidly thrown objects.
\end{abstract}

  
\maketitle

\section{Introduction}
Mobile manipulators (MMs) have emerged as critical tools in diverse domains, including manufacturing, intelligent catering, daily assistance, and medical services \cite{Wang2023, Xie2024, Stibinger2021}. Their combination of enhanced workspace and adaptability makes them particularly suited for complex tasks. Despite these advancements, programming such a high-dimensional robot to operate effectively in dynamic environments remains a persistent challenge. 

For MMs, whole-body planning has been shown to yield smoother and more continuous motions compared to traditional sequential control.  Current whole-body planning methods for MMs are mainly based on Reinforcement Learning (RL) \cite{Wang2020b, Finn2016}, optimization \cite{holistic}, sampling \cite{sampling_based, MotionPlanningRobotics2025}. However, these methods simply implement point-to-point planning of the end-effector, which cannot handle obstacles in the environment during task execution.

\begin{figure}[t]
	\centering
	\includegraphics[width=0.48\textwidth, height=4.5cm]{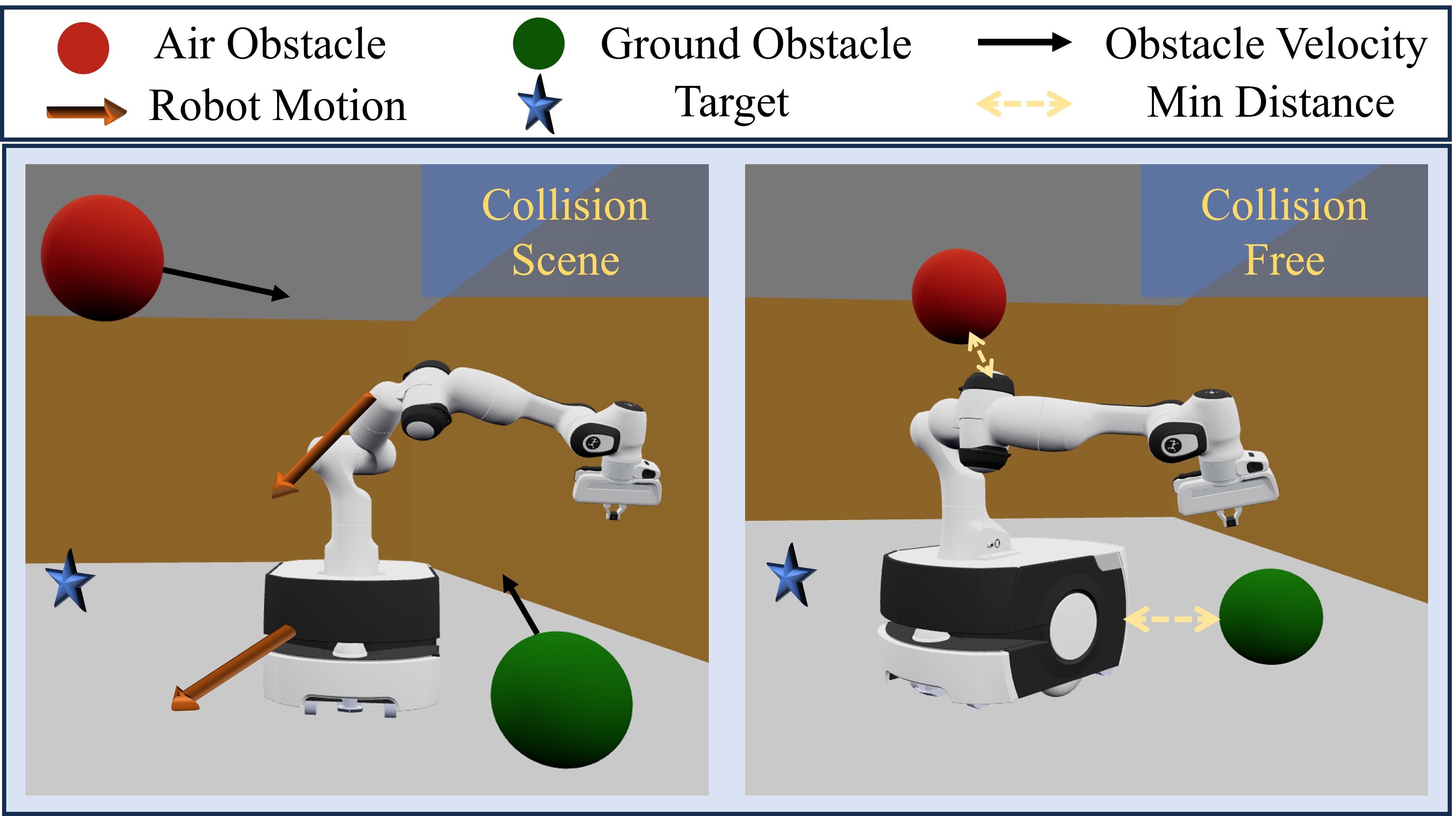}
	\caption{The whole-body motion of the robot when encountering both air and ground dynamic obstacles under SEWB controller.}
	\label{avoid_show}
\end{figure}

Control Barrier Functions  (CBFs) have gained significant traction in ensuring safety by incorporating constraints as linear inequalities \cite{ShawCortez2021, dcbf, Glotfelter2019}. However, their applications have been predominantly limited to low-dimensional spaces. Some work has successfully applied CBF to high-dimensional manipulators \cite{tvcbf, Bertino2022}, focusing on collision avoidance. Nonetheless, prior CBF methods are primarily designed for static obstacles and inability to deal with high-speed obstacles. Moreover, relying on a single CBF constraint can lead to pseudo-equilibrium problems (PEP) \cite{Tan2024, Reis2021, cir}. How to combine CBF to encode more robust safety constraints for MMs remains a significant challenge. 

Accommodate CBF safety constraints in Model Predictive Control (MPC) has shown promise for real-time motion planning of MMs in dynamic environments \cite{mpc_mm, Li2019, Yan2024d}. While MPC can effectively avoid dynamic obstacles, its computational demands grow significantly with increasing symbolic parameters, such as the number of obstacles or prediction horizons. This computational burden often renders MPC impractical for scenarios requiring rapid responses, such as avoiding a flying ball. Conversely, reactive control are naturally suited for high responsiveness and adaptability to environmental changes. Although studies have explored reactive control for avoiding static ground obstacles \cite{EE_T, Ben, neo}, no existing reactive control for MMs can effectively handle both ground and airborne dynamic obstacles. To address these limitations in high-dimensional planning, whole-body control, and responsiveness to dynamic obstacles, this paper proposes a reactive controller designed to achieve safe and expeditious motion for mobile manipulators, as shown in Fig. \ref{avoid_show}.

The primary contributions of this paper are:

1) Create a novel adaptive cyclic inequality  (ACI) approach, combining CBF to establish safety constraints. ACI not only solves the PEP problem associated with conventional CBF-based methods but also enhance the ability to deal with dynamic obstacles. 

2) Propose a safe expeditious whole-body (SEWB) control for mobile manipulators that ensures both external and internal collision-free motion. SEWB can act as a fast-reaction planner to generate whole-body motion to avoid full space obstacles.

3) Simulation and physical experiments, such as avoiding swinging poles and flying balls, are conducted on a 9-DOF mobile manipulator and compared with other works to verify its performance, effectiveness, and stability.

\section{Related Work}
Many MM systems have been designed in recent years. The work in \cite{Fuchs2009} implemented household work like serving tea using an HREB robot which has two seven-axis robotic manipulators. In \cite{Sucan2012}, grasp poses are achieved using an OMPL motion planner which generates motion for the 7 degrees of freedom manipulator to follow. In the works above, the mobile base of the robot is controlled decoupled from the manipulator. This is equivalent to the planning of two unrelated subsystems. Although the method of separate planning simplifies the problem, the task is completed in a slow, discontinuous manner where the motion is stop-start or unnatural.

Speed and gracefulness can be enhanced by treating the mobile base and manipulator as a unified, coordinated control system \cite{Liu2021b, Zhang2022a}. Numerous whole-body planning methods have been proposed to generate trajectories for high degrees of freedom MMs, with a comprehensive review presented in \cite{Sandakalum2022a}. However, these methods of advanced planning are not suitable for dynamic scenarios. Some researchers have introduced a dual trajectory tracking approach for MMs \cite{EE_T, Mashali2018}. However, their method depends on having a fully planned global trajectory and struggles to react to unexpected obstacles. In \cite{holistic}, a reactive whole-body control method based on optimization is proposed, but it completely failed to take obstacles into account. In \cite{Ben}, a reactive approach for MMs is presented, yet it only handles dynamic obstacles on the ground with their motion dependent mainly on the mobile base. These limitations in current approaches highlight the need for controllers that enable MMs to effectively avoid full space fast-moving obstacles.

CBFs have emerged as a prominent method for ensuring safety by maintaining system states within predefined safe sets, formulated as affine inequality constraints on the control input. These constraints, linear in QP-based formulations, rely on local problem information, such as distance function and Lyapunov function gradients, to ensure forward invariance of the safe set \cite{Ames2017}. However, this local approach can result in spurious equilibrium points, where the system may become trapped or even converge to undesired states if initial conditions are unfavorable. To tackle this issue, Reis et al. \cite{Reis2021} proposed an additional CBF constraint to remove boundary equilibrium, while Tan et al. \cite{Tan2024} introduced a modified CBF. In \cite{cir}, a circulation constraint was proposed into the optimization framework. Although these methods mitigate the issue of pseudo-equilibrium, they are ineffective when it comes to handling dynamic obstacles. Moreover, the application of CBF-based methods in high-dimensional systems, such as mobile manipulators, to achieve whole-body obstacle avoidance remains an underexplored area of research.

\section{Modeling and Preliminaries}

\subsection{System Model}
The MM we use includes a mobile base and a manipulator, with $n = n_b + n_m$ degrees of freedom. The state of the manipulator is described by joint positions $\qm\in\mathbb{R}^{n_m}$. For a mobile base with a differential drive, we use the full location description $(x_b,y_b,\varphi)$ and define the $\qb \in \mathbb{R}^{n_b}$ as virtual joints. Therefore, the overall forward kinematics of the robot can be expressed as 
\begin{equation}{}_e^w\boldsymbol{T} = {}_b^w\boldsymbol{T}(x_b,y_b,\varphi) \cdot {}_m^b\boldsymbol{T} \cdot {}_e^m\boldsymbol{T}(\qm),
\end{equation}
where $w$ is the world coordinate system; ${}_b^w\boldsymbol{T}$ is the homogeneous transformation of the mobile base frame relative to the world coordinate system. ${}_m^b\boldsymbol{T}$ is a constant relative pose from the mobile base frame to the manipulator frame and ${}_e^m\boldsymbol{T}$ is the position forward kinematics of the manipulator where the end-effector (EE) frame is $e$.

Similar to the Jacobian matrix in the velocity forward kinematics of the manipulator, we define the extended Jacobian matrix ${}_e^w\boldsymbol{J} \in  \mathbb{R}^{6\times n}$ of the MM to represent the EE's velocity in the world. Therefore, we can obtain the velocity forward kinematics of the MM as follows:
\begin{equation}
	\label{exp_v}
	{}_e^w\boldsymbol{v} = {}_e^w\boldsymbol{J}(x_b,y_b,\varphi ,\qm)\dot{\boldsymbol{q}},
\end{equation}
where $\dot{\boldsymbol{q}}=(\dot{\boldsymbol{q}}_b,\dot{\boldsymbol{q}}_m)^\text{T}$. This maps the velocity of all axes of the MM to the EE velocity ${}_e^w\boldsymbol{v}=(v_x,v_y,v_z,\omega_x,\omega_y,\omega_z)^\text{T}$.

\subsection{Primary Optimization}
In general, we formulate a quadratic programming (QP) problem with inequality constraints to achieve expeditious reactive control.
\begin{gather}
    \boldsymbol{u}^*(\q) = \arg\min_{\boldsymbol{u}} \|\boldsymbol{u}-\boldsymbol{u}_d(\q) \|^2 \label{QP}  \\
	\text{s.t.} \quad b_i(\boldsymbol{u}) \geq \boldsymbol{p}_i,\quad i = 1, 2\dots \label{inequa}
\end{gather}
where $\boldsymbol u$ is the system input. The QP aims to make the smallest possible adjustments to the nominal control $\boldsymbol{u}_d(\q)$, while ensuring that the constraints are satisfied. We use $b_i(\boldsymbol{u}) \geq \boldsymbol{p}_i $ to uniformly represent all inequality constraints, and the specific forms of these constraints will be introduced in detail later.

\begin{figure}[t]
	\centering
	\includegraphics[height=5.5cm, width=0.3\textwidth]{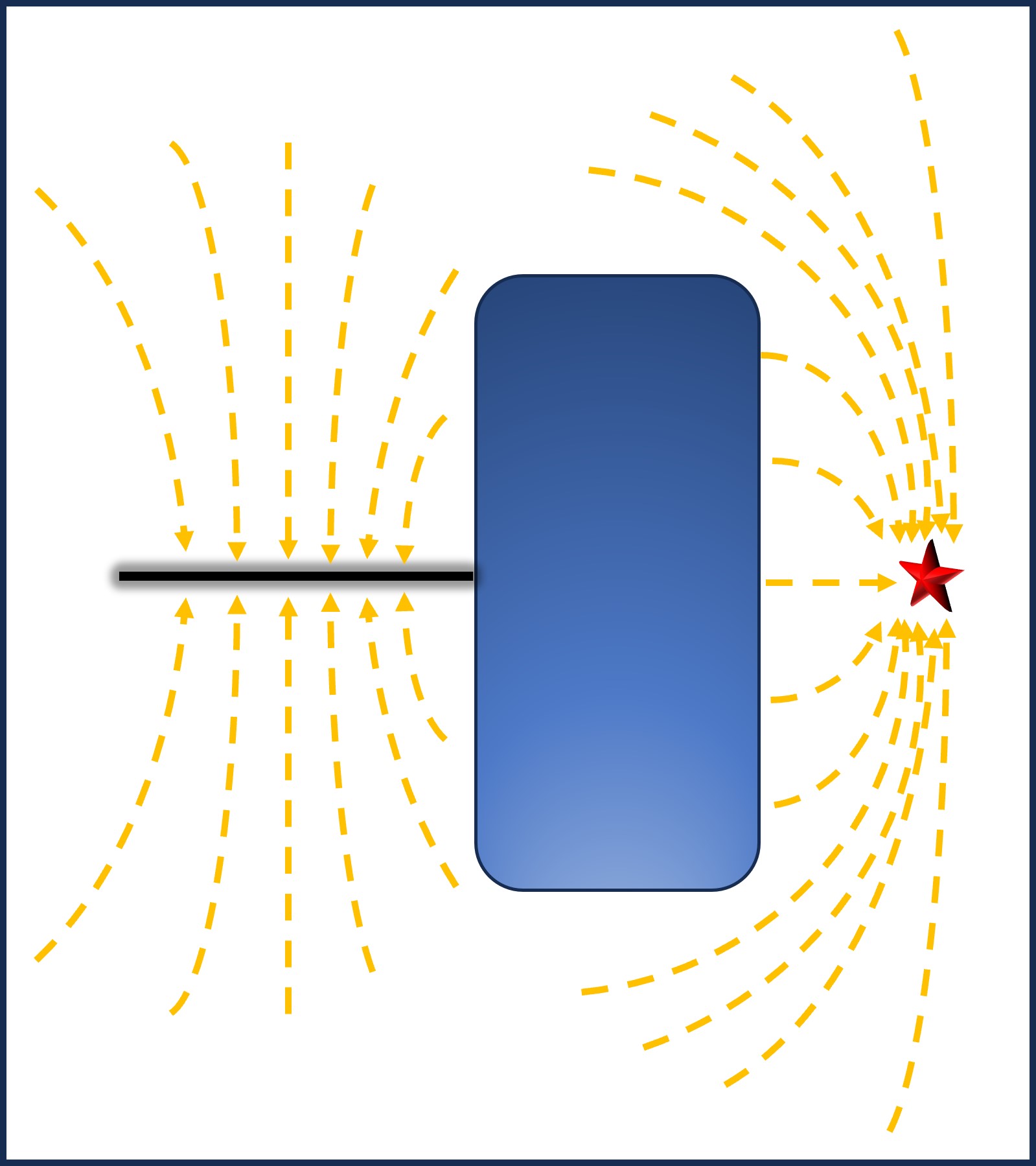}
	\caption{The vector field around the obstacle after applying CBF-QP, where pseudo-equilibrium points appear along the black line. }
	\label{cbf_pep}
\end{figure}

\subsection{Control Barrier Functions}
CBFs ensure safety by leveraging the concept of forward invariance within a designated safe set, defined as:
\begin{equation}
\label{C_}
\mathcal{C}=\left\{\q \in \mathbb{R}^{n} \mid h(\q) \geq 0\right\},
\end{equation}
where $\mathcal{C}$ denotes the interior of the security set. The function $h:\mathbb{R}^n\to\mathbb{R}$  represents a measure of "closeness" between the robot to obstacle \cite{cir}, and it should be continuously differentiable. Then, if there exists a class $\mathcal{K}$ function $\alpha$, such as $\alpha(x) = -ax$ for a constant $a>0$, the CBF constraint can be defined as follows,
\begin{equation}
	\label{cbf}
    \exists \boldsymbol{u} \quad s.t. \quad \dot{h}(\q) \cdot \boldsymbol{u} \geq \alpha(h(\q)), 
\end{equation}
As a result of the CBF theorem,  $\mathcal{C}$ is forward invariance \cite{9561253, 9812378}. However, when CBF constraints are integrated into a QP framework, the issue of pseudo-equilibrium points can arise \cite{cir, Singletary2021}. 

We consider a minimally invasive control strategy defined by the QP Eq.(\ref{QP}), subject to the safety constraint $\dot{h}(\q)\boldsymbol{u} \geq \alpha(h(\q))$. This optimization is strictly convex and always has a unique solution if the constraint is feasible. Using the Karush-Kuhn-Tucker (KKT) conditions, the optimal solution can be expressed as
\begin{gather}
    \boldsymbol{u}^*(\q) = \boldsymbol{u}_d(\q) + \lambda(\q) \nabla h(\q),  \\
    \lambda(\q) = \max\left(0, -\frac{\dot{h}(\q) \boldsymbol{u}_d(\q) + \alpha(h(\q))}{\|\nabla h(\q)\|^2}\right),
\end{gather}
where $\lambda(\q)$ is the nonnegative Lagrange multiplier associated with the CBF constraint. This expression shows that the optimal control consists of the nominal input $\boldsymbol{u}_d(\q)$ plus a configuration-dependent repulsive term that becomes active only when the system approaches the boundary of the safe set.
However, this repulsive term may counteract the nominal control in certain configurations. Specifically, when the nominal control $\boldsymbol{u}_d(\q)$ tends to violate the safety constraint, the barrier term $\lambda(\q)\nabla h(\q)$ opposes it to enforce safety. This opposition can cause the two terms to cancel each other out exactly, resulting in $\boldsymbol{u}^*(\q) = 0$ even though the system has not reached its goal. Such undesired stationary points are known as pseudo-equilibria, , as shown in Fig. \ref{cbf_pep}. They arise due to the conflict between pursuing the nominal objective and maintaining safety, and are a common issue in barrier-based control strategies.

\section{Control Methodology}
\subsection{Adaptive Cyclic Inequality}
\begin{figure}[t]
	\centering
	\includegraphics[width=0.48\textwidth, height=4cm]{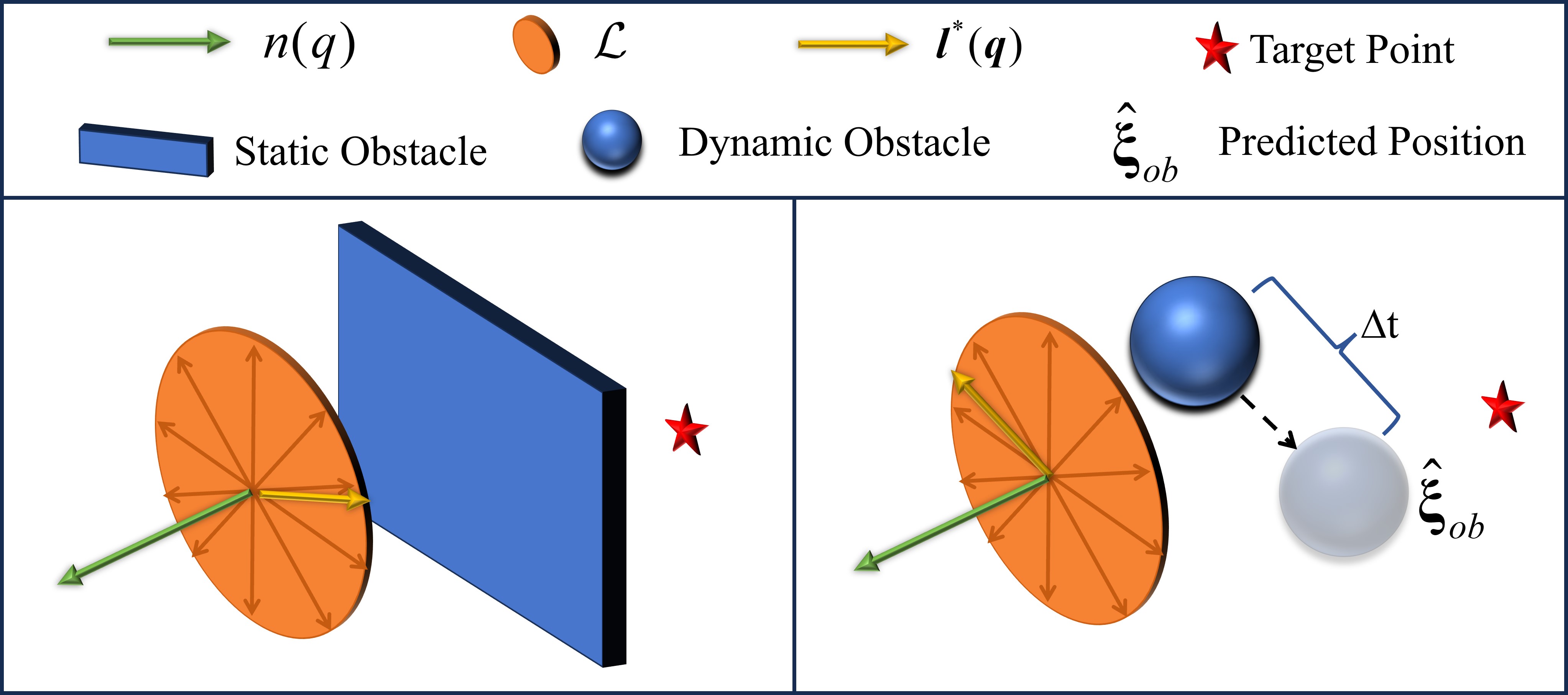}
	\caption{The left figure illustrates an intuitive diagram of the variables used in ACI computation when encountering a static obstacle in three-dimensional space. As shown,  $\mathcal{L}$ is a set containing infinitely many vectors, and the goal of ACI is to select $\boldsymbol{l}^*(\q)$ from this set to be used as the input for Eq.(\ref{aci}). The right figure shows the variable illustration in the case of a dynamic obstacle. The future positions of the dynamic obstacle are predicted and used as the basis to compute the optimal $\boldsymbol{l}^*(\q)$. In a real robot system, ACI operates in a higher-dimensional space (e.g, $n=7$), but three-dimensional space is used here for ease of visualization.}
	\label{aci_show}
\end{figure}

In addition to the issue of pseudo-equilibrium points, the function $h(x)$ in the CBF is usually designed for static obstacles and fixed safety boundaries, which makes it hard to update in real-time for dynamic obstacles. When obstacles move or change position, the $h(x)$ function may not reflect these changes, leading to poor responsiveness. To address this, we enhance the CBF with an adaptive cyclic inequality (ACI) constraint that encourages the system to move around obstacles. To begin, we define the normal vector,
\begin{equation}
	n(\q) = \nabla h(\q) / \|\nabla h(\q) \|.
\end{equation}
Then we can define a set of unit vectors that:
\begin{equation}
	\mathcal{L} =\left\{ \boldsymbol{l}(\q) \mid \boldsymbol{l}(\q) \cdot n(\q) = 0\right\}, \label{L}
\end{equation}
which represents the set of all unit orthogonal vectors for $n(\q)$. When $n(\q) \in \mathbb{R}^{2}$, there are exactly two unit vectors in $\mathcal{L}$ that are perpendicular to $n(\q)$. Meanwhile, if $n(\q) \in \mathbb{R}^{3}$ or a higher dimension, it is obvious that $\mathcal{L}$ will have an infinite number of vectors. Therefore, in any case, $\mathcal{L}$ is a multi-solution set. A very natural and meaningful idea is to select the most favorable solution from this set under a certain cost condition. This process is exactly what the "Adaptive" in ACI refers to. We first define the optimal solution in $\mathcal{L}$ as $\boldsymbol{\l}^*(\q)$. During the process of obtaining $\boldsymbol{l}^*(\q)$ in $\mathcal{L}$, specifically, we hope to enhance the ability to better avoid dynamic obstacles and to better track nominal control. With this, the focal point is updated as follows,
\begin{gather}
	\label{sub-opti}
	\boldsymbol{l}^*(\q) = \max_{\boldsymbol{l}} M(\boldsymbol{l}, \hat{\boldsymbol{\xi}}_{ob}) \quad\forall  \boldsymbol{l} \in \mathcal{L}, \\
	\label{M}
	M(\boldsymbol{l}, \hat{\boldsymbol{\xi}}_{ob}) =  \zeta \boldsymbol{l} \cdot \nabla{h}(\q, \hat{\boldsymbol{\xi}}_{ob}) + (1-\zeta) \boldsymbol{l} \cdot \boldsymbol{u}_d(\q).
\end{gather}
The term $\hat{\boldsymbol{\xi}}_{ob}$ denotes the predicted future state of obstacles after $\Delta t$, while $\nabla h(\q, \hat{\boldsymbol{\xi}}_{ob})$ represents the predicted gradient of the function $h$ with respect to the predicted future obstacle configuration.  The adaptive function $M$ characterizes the degree of directional alignment among $\boldsymbol{l}$, $\nabla h(\q, \hat{\boldsymbol{\xi}}_{ob})$ and $\boldsymbol{u}_d(\q)$. To make $M$ larger, the term $\boldsymbol{l} \cdot \nabla h(\q, \hat{\boldsymbol{\xi}}_{ob})$ increases the projection of $\boldsymbol{l}$ onto $\nabla h(\q, \hat{\boldsymbol{\xi}}_{ob})$, thereby guiding the trajectory away from predicted obstacle locations. Similarly, $\boldsymbol{l} \cdot \boldsymbol{u}_d(\q)$ also increases the projection, make $\boldsymbol{l}$ tends to approach the direction of the nominal control $\boldsymbol{u}_d(\q)$. $\zeta \in (0,1)$ is a proportion coefficient, representing the weight of the two items. A visual interpretation of this function is that the selected direction  $\boldsymbol{l^*}(\q)$ tends to increase the distance from future obstacles and better align with the direction toward the target. Then, we present the constraint form of ACI as follows,
\begin{equation}
    \label{aci}
    \exists \boldsymbol{u} \quad s.t. \quad \boldsymbol{l}^*(\q) \cdot \boldsymbol{u} \geq T(h(\q)),
\end{equation}
where $T$ is a continuous function: $T:\mathbb{R}^+ \rightarrow \mathbb{R}$ with the decreasing properties. $T(0)>0$, $T(h) \rightarrow -\infty$ when $h \rightarrow +\infty$. The constraint threshold function $T$ facilitates ACI gradually increase its directional effect on $\boldsymbol{u}$ as it nears an obstacle. Fig. \ref{aci_show} shows the  $\boldsymbol{l^*}(\q)$ of ACI in three-dimensional space for both static and dynamic obstacles.

\subsection{Equilibrium Points Analysis}
We consider the optimization problem~Eq.(\ref{QP}) with the CBF constraint~Eq.(\ref{cbf}) and the ACI constraint~Eq.(\ref{aci}). Our goal is to characterize the set of configurations $\boldsymbol{q}$ for which the optimal control input vanishes, i.e., $\boldsymbol{u}^*(\boldsymbol{q}) = \boldsymbol{0}$, corresponding to equilibrium points of the closed-loop system $\dot{\boldsymbol{q}} = \boldsymbol{u}^*(\boldsymbol{q})$.

To this end, we analyze the Karush-Kuhn-Tucker (KKT) conditions associated with the QP. Let $\lambda_h, \lambda_T \in \mathbb{R}$ be the dual variables associated with the CBF and ACI constraints, respectively. When $\boldsymbol{u}^* = \boldsymbol{0}$ is optimal, the KKT conditions reduce to:
\begin{equation}
\begin{aligned}
\text{(i)} &\quad \boldsymbol{u}_d + \lambda_h \nabla h + \lambda_T \boldsymbol{l}^* = \boldsymbol{0}, \\
\text{(ii)} &\quad \lambda_h \left( \nabla h^\top \boldsymbol{u} + \alpha(h(\boldsymbol{q})) \right) = 0, \\
\text{(iii)} &\quad \lambda_T \left( \boldsymbol{l}^{*\top} \boldsymbol{u} + T(h(\boldsymbol{q})) \right) = 0, \\
\text{(iv)} &\quad \lambda_h \geq 0, \quad \lambda_T \geq 0, \\
\text{(v)} &\quad \nabla h \cdot \boldsymbol{u} \geq \alpha(h(\boldsymbol{q})), \\
\text{(vi)} &\quad \boldsymbol{l}^* \cdot \boldsymbol{u} \geq T(h(\boldsymbol{q})).
\end{aligned}
\end{equation}
Since $\boldsymbol{u} = \boldsymbol{0}$, conditions (v) and (vi) reduce to:
\begin{equation}
\alpha(h(\boldsymbol{q})) \leq 0, \quad T(h(\boldsymbol{q})) \leq 0.
\end{equation}
Suppose $\alpha(h(\boldsymbol{q})) \neq 0$ (i.e., safety is guaranteed). Then (ii) implies $\lambda_h = 0$. Thus, the stationarity condition (i) becomes as follow:
\begin{equation}
    \boldsymbol{u}_d + \lambda_T \boldsymbol{l}^*(\boldsymbol{q}) = \boldsymbol{0}.
\end{equation}

By analyzing \text{(iii)}, we can conclude that either $\lambda_T=0$ or $ T(h(\boldsymbol{q}))=0$. Two mutually exclusive cases arise, corresponding to different physical interpretations:

\textbf{Case 1:} If $\lambda_T = 0$, then $\boldsymbol{u}_d = \boldsymbol{0}$, meaning that the desired control input vanishes. This implies the system has reached the goal and no further motion is needed.

\textbf{Case 2 :} If $T(h(\boldsymbol{q})) = 0$ and $\lambda_T > 0$, then $\boldsymbol{u}_d$ is negatively aligned with $\boldsymbol{l}^*(\boldsymbol{q})$, that is, $\boldsymbol{l}^*(\boldsymbol{q}) = -\frac{1}{\lambda_T} \boldsymbol{u}_d = -\lambda \boldsymbol{u}_d$ for some $\lambda > 0$. This situation corresponds to a pseudo-equilibrium point under the CBF-ACI constraints, where the circulation constraint exactly cancels the desired motion. However, in the formulation of~Eq.(\ref{M}), we explicitly select $\boldsymbol{l}^*(\boldsymbol{q})$ to be positively aligned with the direction of $\boldsymbol{u}_d$, and the negative aligned is the least preferred. As a result, such pseudo-equilibrium points are effectively avoided.

\subsection{Feasibility Analysis}

When there are obstacles, one important question is whether CBF and ACI as inequality constraints will conflict with each other. The following result answers this question positively.

\textit{Theorem 1: } Let $\beta$ be such that $T(0) < \beta < r$, where $r$ is a known control limit (e.g., a bound on the norm of the velocity command). Then, when the system is in the safe set $\mathcal{C}$, that is $h(\boldsymbol{q}) \geq 0$ and $\alpha(h(\boldsymbol{q})) \leq 0$, the QP formulation with constraints Eq.(\ref{cbf}), Eq.(\ref{aci}) and $\|\boldsymbol{u}\| \leq r$ admits a feasible solution:  
\begin{equation}
  \boldsymbol{u} = \beta \boldsymbol{l}^*(\boldsymbol{q}).  
\end{equation}

\textit{Proof: }We shall verify all constraints in Eq.(\ref{cbf}), Eq.(\ref{aci}), and the norm bound $\|\boldsymbol{u}\| \leq r$.  
\begin{enumerate}
    \item \textbf{CBF constraint Eq.(\ref{cbf}):}  
    Note that $\dot{h}(\boldsymbol{q}) \cdot \boldsymbol{u} =  \beta \dot{h}(\boldsymbol{q})^\top  \allowbreak \boldsymbol{l}^*(\boldsymbol{q})$.  
    From the definition of $\boldsymbol{l}^*(\boldsymbol{q})$ in Eq.(\ref{L}), we observe that $\dot{h}(\boldsymbol{q})^\top \boldsymbol{l}^*(\boldsymbol{q}) = 0$.  
    Hence, $\dot{h}(\boldsymbol{q}) \cdot \boldsymbol{u} = 0 \geq \alpha(h(\boldsymbol{q}))$, which holds because $\alpha(h(\boldsymbol{q})) \leq 0$.

    \item \textbf{ACI constraint Eq.(\ref{aci}):}  
    We compute $\boldsymbol{l}^*(\boldsymbol{q}) \cdot \boldsymbol{u} = \beta{\boldsymbol{l}^*(\boldsymbol{q})}^\top \boldsymbol{l}^*(\boldsymbol{q}) = \beta$, since $\boldsymbol{l}^*(\boldsymbol{q})$ is a unit vector and therefore satisfies ${\boldsymbol{l}^*(\boldsymbol{q})}^\top \boldsymbol{l}^*(\boldsymbol{q}) = 1$. Moreover, because $T$ is a decreasing function, its maximum value occurs at $h = 0$, and hence $\beta > T(0) \geq T(h(\boldsymbol{q}))$. This confirms that the ACI constraint is satisfied.

    \item \textbf{Norm bound constraint:}  
    The norm of the control input is $\|\boldsymbol{u}\| = \|\beta \boldsymbol{l}^*(\boldsymbol{q})\| = \beta$, since $\|\boldsymbol{l}^*(\boldsymbol{q})\| = 1$.  
    Because $\beta < r$ by assumption, we have $\|\boldsymbol{u}\| < r$, so the constraint is strictly satisfied.
\end{enumerate}

Thus, the constraints do not conflict with each other and $\boldsymbol{u} = \beta \boldsymbol{l}^*(\boldsymbol{q})$ lies in the interior of the feasible set.


\section{Application for Mobile Manipulators}
The whole-body reactive collision avoidance for mobile manipulators is challenging because it consists of a low-dimensional mobile base and a high-dimensional manipulator. We are considering applying the obstacle avoidance algorithm to both parts simultaneously to verify the universality. 

\begin{figure}[t]
	\centering
	\includegraphics[width=0.48\textwidth]{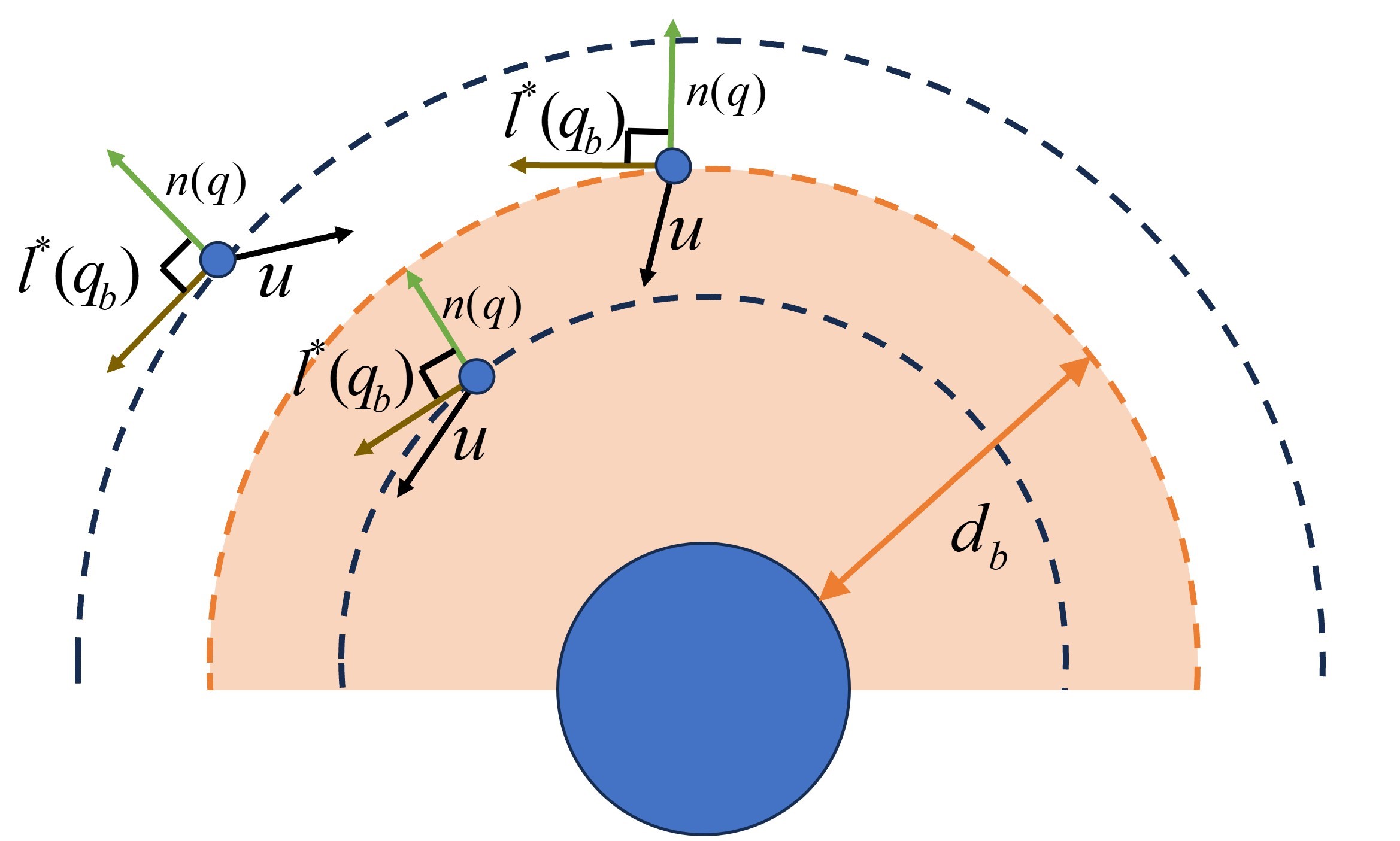}
	\caption{The analysis of ACI constraint for mobile base in two-dimensional space. As the distance to the obstacle decreases, the final generated $\qbd$ increasingly aligns with $\boldsymbol{l}^*(\qb)$. The threshold distance $d_b$ determines the range within which the ACI exerts a positive influence.}
	\label{Tb_show}
\end{figure}

\subsection{External Safety Constraints}
\textit{1) Mobile Base:} Since the mobile base can be represented by a single shape, we directly use the Euclidean distance to define its safety function $h$ as $h_b$. In ACI constraint~Eq.(\ref{aci}), we denote the transformation $T$, $\boldsymbol{l}^*(\q)$ associated with the mobile base as $T_b$ and $\boldsymbol{l}^*(\qb)$, respectively. The mobile base's configuration is represented by the joint variable $\qb$. The resulting formulation of $T_b$ is given by:
\begin{equation}
    T_b = d_b - h_b(\qb), \label{d_b}
\end{equation}
where $d_{b}$ represents the threshold parameter for ACI to initiate a positive constraint on mobile base. When $d_b - h_b(\qb) > 0 $, it means that ACI has turned on the positive constraint on the control variable $\boldsymbol{u}$. Fig. \ref{Tb_show} shows the gradual enhancement of ACI constraint for mobile base in a two-dimensional space.

\textit{2) Manipulator:} The manipulator is a high-dimensional system with a complex rigid-body structure, making the computation of its safety function $h$ more involved than that of the mobile base. Specifically, its collision model comprises $m$ primitive shapes and is actuated by $\qm$ joints. The overall safety set for the manipulator is defined by requiring that each individual shape satisfies its respective safety condition, expressed as:
\begin{equation}
    \mathcal{C}_{m}=\left\{\boldsymbol{q}_{m} \mid \mathcal{G}_{i}\left(\boldsymbol{q}_{m}\right) \geq 0, \quad i=1,2, \ldots, m\right\},
\end{equation}
The functions $\mathcal{G}_{i}$ represent the safety conditions for each shape, typically derived by computing the distance between two rigid objects, each modeled as a configuration-dependent 3D convex set. For example, to compute the distance between a manipulator link and an obstacle, algorithms such as von Neumann’s cyclic projection algorithm \cite{Bauschke1993} are used to find the closest points, denoted by $\boldsymbol{\rho}_m$ on the manipulator and $\boldsymbol{\rho}_o$ on the obstacle. In this case, the safety function $\mathcal{G}_{i}$ is defined as half the squared distance between these closest points.
\begin{equation}
\mathcal{G}_{i}\left(\boldsymbol{q}_{m}\right)=\frac{1}{2}\left\|\boldsymbol{\rho}_{m}-\boldsymbol{p}_{o}\right\|^{2}.
\end{equation}
Then, the gradient of $\mathcal{G}_{i}$ with respect to the manipulator’s configuration $\qm$ can be computed numerically, for example, using finite difference methods. Regardless of how the functions $G_i$ are obtained, the overall safety function $h$ can be defined as the minimum of these $\mathcal{G}_{i}$ functions \cite{Glotfelter2017}. However, even if each $\mathcal{G}_{i}$ is differentiable, the function $h$ may exhibit non-differentiable points due to the nature of the minimum operation \cite{cir}. To address this, a smoother approximation such as the softmin \cite{Gao2017a} can be used:
\begin{equation}
\mathrm{softmin}_i^z(\mathcal{G}_i)=z\ln\left(\frac{1}{m}\sum_{i=1}^{m}\exp\left(-\frac{\mathcal{G}_i}{z}\right)\right),
\end{equation}
where $z$ is a smoothing parameter. The softmin function provides a differentiable approximation but satisfies $\mathrm{softmin}_z^i(\mathcal{G}_i)  \allowbreak \geq \min_i(\mathcal{G}_i)$. Consequently, $\mathrm{softmin}_z^i(\mathcal{G}_i) > 0$ does not necessarily imply that $\mathcal{G}_i > 0$ for all $i$. To ensure that $h$ remains a valid safety function, it is beneficial to introduce a margin $\delta_s$. We then define the safety function $h$ for the manipulator $h_m$ as:
\begin{equation}
    h_m(\boldsymbol{q}_m)=\mathrm{softmin}_i^z(\mathcal{G}_i)-\delta_s.
\end{equation}
This modification preserves the desired properties of $h_m$ while benefiting from the smoothness of the softmin approximation. 

Then, in the ACI constraint Eq.(\ref{aci}), we denote the transformation $T$ associated with the manipulator as $T_m$, defined as follows:
\begin{equation}
T_m = d_m - h_m(  \qm ), \label{d_m}
\end{equation}
where $d_{m}$ represents the threshold parameter for ACI to initiate a positive constraint on manipulator. When $d_m - h_m(\qm) > 0 $, it means that ACI has turned on the positive. The principle is similar to that expressed in Fig. \ref{Tb_show}, but in a higher dimensional space.

As a result, the inequality constraints in Eq.(\ref{inequa}) for external safety are as follows:
\begin{equation}
\begin{aligned}
&(\dot{h}_b(\qb), \boldsymbol{0}) \cdot \boldsymbol{u} \geq \alpha(h_b(\qb)), \\
&( \boldsymbol{l}^*(\qb), \boldsymbol{0}) \cdot \boldsymbol{u} \geq T_b(h_b(\qb)), \\
&(\boldsymbol{0}, \dot{h}_m(\qm)) \cdot \boldsymbol{u} \geq \alpha(h_m(\qm)), \\
&( \boldsymbol{0}, \boldsymbol{l}^*(\qm)) \cdot \boldsymbol{u} \geq T_m(h_m(\qm)).
\end{aligned}
\end{equation}

\begin{figure}[t]
	\centering
	\includegraphics[width=0.48\textwidth]{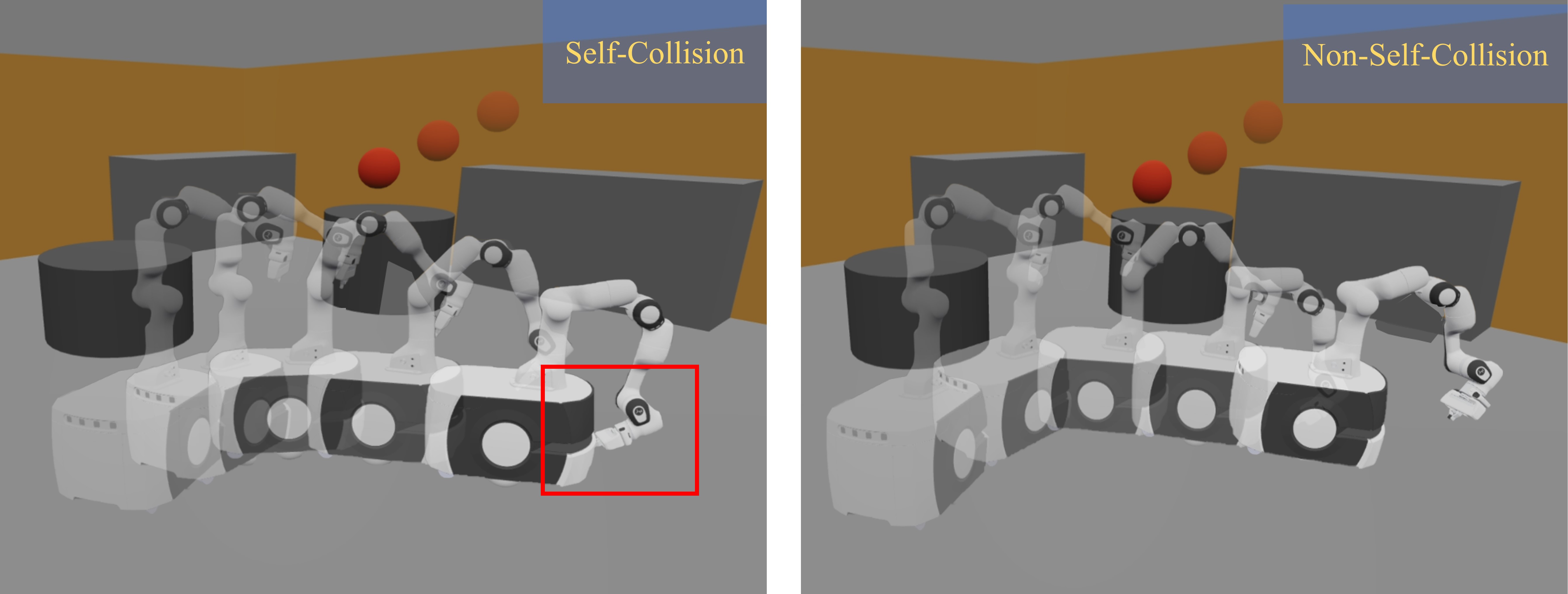}
	\caption{Before adding the self collision constraint in internal safety, the manipulator encountered collision with the base. Incorporating the constraint Eq.(\ref{sc}) effectively resolved this issue.}
	\label{self_collision}
\end{figure}

\subsection{Internal Safety Constraints}
In the previous subsection, a combination of ACI and CBF was introduced to guarantee external safety. In addition, the internal safety of the system in the process of movement also needs to be strictly guaranteed \cite{sampling_based}. Internal safety includes joint limit enforcement and self-collision avoidance (particularly between the mobile base and the manipulator). For the joint limit enforcement of the manipulator, a simple function of joint bounds can be derived for each joint to keep it between its lower $q_i^-$ and upper bounds $q_i^+$, where $i \in [1, q_m]$.
\begin{equation}
	b_{jl}^i = \frac{(q_i^+-q_i)(q_i-q_i^-)}{q_i^+-q_i^-}.
\end{equation}
To address self-collision between the mobile base and the manipulator, we introduce a range limit on the manipulator’s reach relative to the base. The following expression defines a valid CBF for enforcing this reach constraint.
\begin{equation}
    b_{sc}=\left(p_{\max}^2-(\boldsymbol{p_e}-\boldsymbol{p_b})^TP(\boldsymbol{p_e}-\boldsymbol{p_b})\right).
\end{equation}
where $\boldsymbol{p_e}$ and $\boldsymbol{p_b}$ denote the positions of the end-effector and the base, respectively. The scalar $p_{\max}$ specifies the maximum allowable reach. The matrix $P = \text{diag}(1, 1, 0)$ is a diagonal selector that preserves only the $x$ and $y$ components, ensuring that the reach is evaluated within the 2D plane. Based on this, the CBF constraint for internal safety in Eq.(\ref{inequa}) are formulated as follows:
\begin{align}
	\label{jl}
	\dot{b}_{jl}^{i} \cdot u &\geq-\gamma_{jl} b_{jl}\quad\forall i\leq q_m \quad \text{(Joint Limits)}\\
	\label{sc}
	\dot{b}_{sc} \cdot u &\geq-\gamma_{sc} b_{sc} \quad \text{(Self Collision)}. 
\end{align}

Fig. \ref{self_collision} illustrates the role of internal safety constraints in preventing robot self-collisions, particularly in avoiding collisions between the mobile base and the manipulator. 

\subsection{Whole-body Nominal Control}
In the primary optimization Eq.(\ref{QP}), a nominal control input $\boldsymbol{u}_d(\q)$ needs to be computed in advance. Consider the whole-body kinematic model of the mobile manipulator Eq.(\ref{exp_v}), we formulate the following optimization problem to map the expected EE velocity ${}_e^w\boldsymbol{v}^*(t)$ to the corresponding expected joint velocity $\boldsymbol{u}_d(\boldsymbol{q})$.
\begin{gather}
\boldsymbol{\nu}^* = \min_{\boldsymbol{\nu}} \frac{1}{2}\boldsymbol{\nu}^\text{T}\boldsymbol{H}\boldsymbol{\nu} + \boldsymbol{g}^\text{T}\boldsymbol{\nu}	\label{QP_v}\\
\text{s.t.} \quad {}_e^w\boldsymbol{v}(t) + \boldsymbol{\delta}(t) = {}_e^w\boldsymbol{v}^*(t), \label{equa}
\end{gather}
where $\boldsymbol{\nu} = (\qd, \boldsymbol{\delta})^\text{T}$ is the decision variable and $\boldsymbol{\delta} \in  \mathbb{R}^6$ is the slack vector. ${}_e^w\boldsymbol{v}^*(t)$ can be obtained through proportional control of the difference between the desired EE position and the current position. In the cost function, $\boldsymbol{H}$ and $\boldsymbol{g}$ represent the cost function of quadratic and linear coefficients, respectively.  Similar in \cite{holistic}, $\boldsymbol{H}$ is a positive definite diagonal matrix used to adjust the weight of each optimization variable; $\boldsymbol{g}$ is designed to maximize the manipulability of the manipulator and adjust the direction of both the mobile base and the manipulator. Finally, the $\boldsymbol{u}_d(\q)$ = $\qd$ in $\boldsymbol{\nu}$. 

\begin{figure}[t]
	\centering
	\includegraphics[width=0.48\textwidth, height=5.5cm]{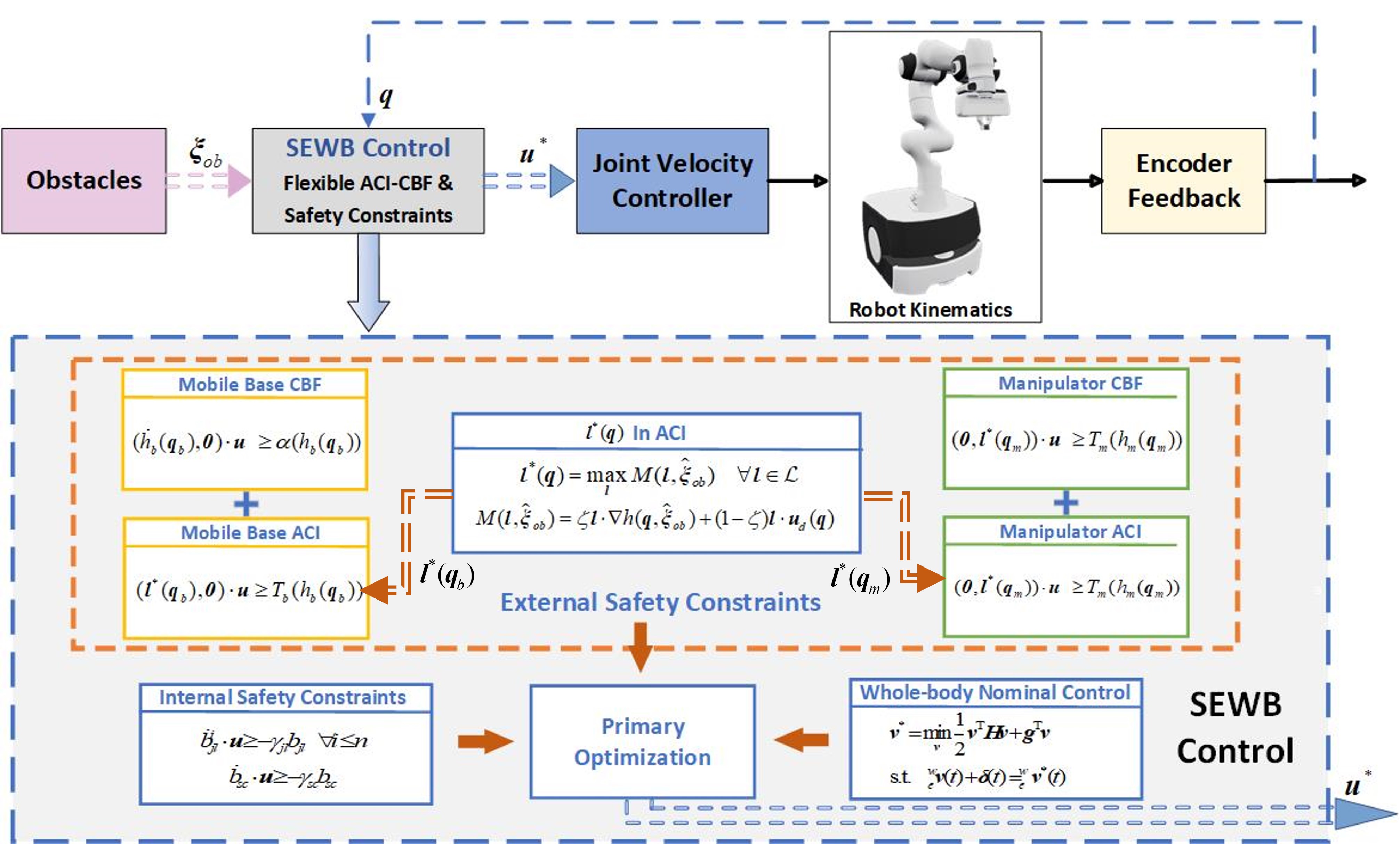}
	\caption{Framework of the SEWB control. ACI and CBF together form the external safety constraints of the main optimization, which, along with internal safety constraints, constitute the whole-body safety control of the mobile manipulator. Finally, the main optimization outputs the desired joint velocities to the robot controller.}
	\label{frame}
\end{figure}

Fig. \ref{frame} presents our entire control framework of SEWB.

\begin{figure*}[t]
	\centering
	\includegraphics[width=\textwidth,height=3cm]{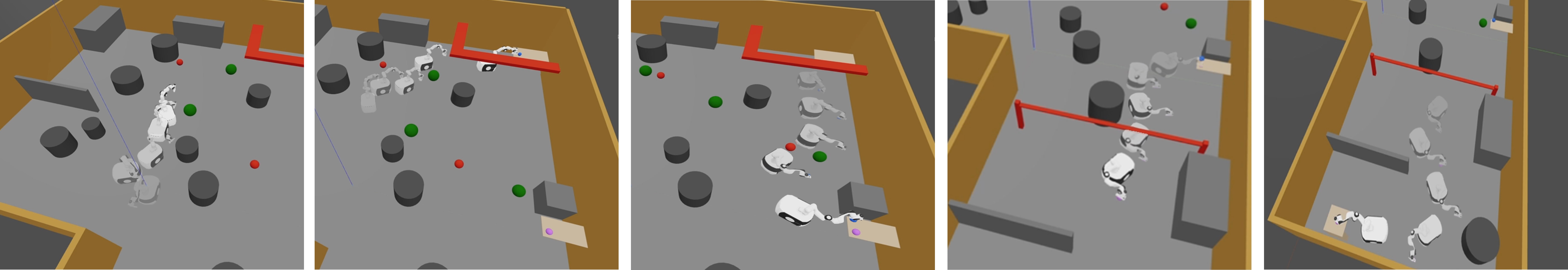}
	\caption{Snapshots of the mobile manipulator's motion in Experiment 1 using SEWB. The timestamp of the snapshots increases from left to right.}
	
	\label{exp1_show}
\end{figure*}

\section{Experiments}
We evaluate our approach through experiments conducted both in simulation and on a real mobile manipulator. We use the following values for the parameters of the controller: $\zeta=0.8$ in Eq.(\ref{M}); $d_b=0.3$ in Eq.(\ref{d_b}); $d_m=0.25$ in Eq.(\ref{d_m}); $\gamma_{jb}=0.1 $ in Eq.(\ref{jl}) and $\gamma_{mr}=0.1$ in Eq.(\ref{sc}). The two important parameters of our controller are $d_{b} $ and $ d_{m}$, which we will discuss about the effects of changing them in Experiment 1. We employ a second-order Kalman filter that accounts for acceleration to predict the future position of the obstacle in~Eq.(\ref{M}), setting $\Delta t = 0.25s$. In calculating the safety functions $h_b$ and $h_m$, we used the Python roboticstoolbox \cite{rtb}. Both the mobile base and the manipulator links have cylindrical collision models defined in the URDF. For the simulated experiments, we use the Swift and Python roboticstoolbox \cite{rtb}.  For the physical experiments, we use a Sunspeed IR-C100 mobile base ($n_b=2$) and a Franka Panda manipulator ($n_m=7$), interfacing with the robot through ROS Noetic. We use a nokov motion capture to acquire the state information of the environment and robot. The QP problems are solved using the DAQP package. In Eq.(\ref{sub-opti}), We use a finite sampling approach to compute a limited number of vectors in set $\mathcal{L}$, from which the optimal vector is selected. All experiments are performed on a laptop with 32GB of RAM and an i9-14900HX processor.

\subsection{Experiment 1: Simulation Scenario}
The simulation presents a dynamic scenario where the MM sequentially moves to the target positions \((-6, 4, 0.68)\), \((-0.1, 4.3, 0.78)\), \((0.1, 4.3, 0.76)\), and \((6.6, 1, 0.76)\). Figure \ref{exp1_show} illustrates the motion of the MM using our controller. In the figure, the red ball represents a dynamic obstacle at a certain height, while the green ball represents a dynamic obstacle on the ground. Both obstacles move at a speed of 2m/s. Dynamic obstacles in the environment are modeled using spherical collision geometries, while static obstacles are modeled using cylindrical or box-shaped collision models. We compare our controller with other existing methods that solve the pseudo-equilibrium problem of CBF \cite{Reis2021,cir} and the MPC-based method \cite{mpc_mm}. Additionally,  we conducted an ablation study comparing several configurations: CBF alone, CBF+ACI (b)  (ACI for the mobile base), and CBF+ACI (b,m)  (ACI for both the mobile base and manipulator). Furthermore, we evaluate the performance of our controller by varying the parameters $d_b$ and $d_m$.

\begin{figure}[t]
	\centering
	\includegraphics[width=0.48\textwidth,height=3.5cm]{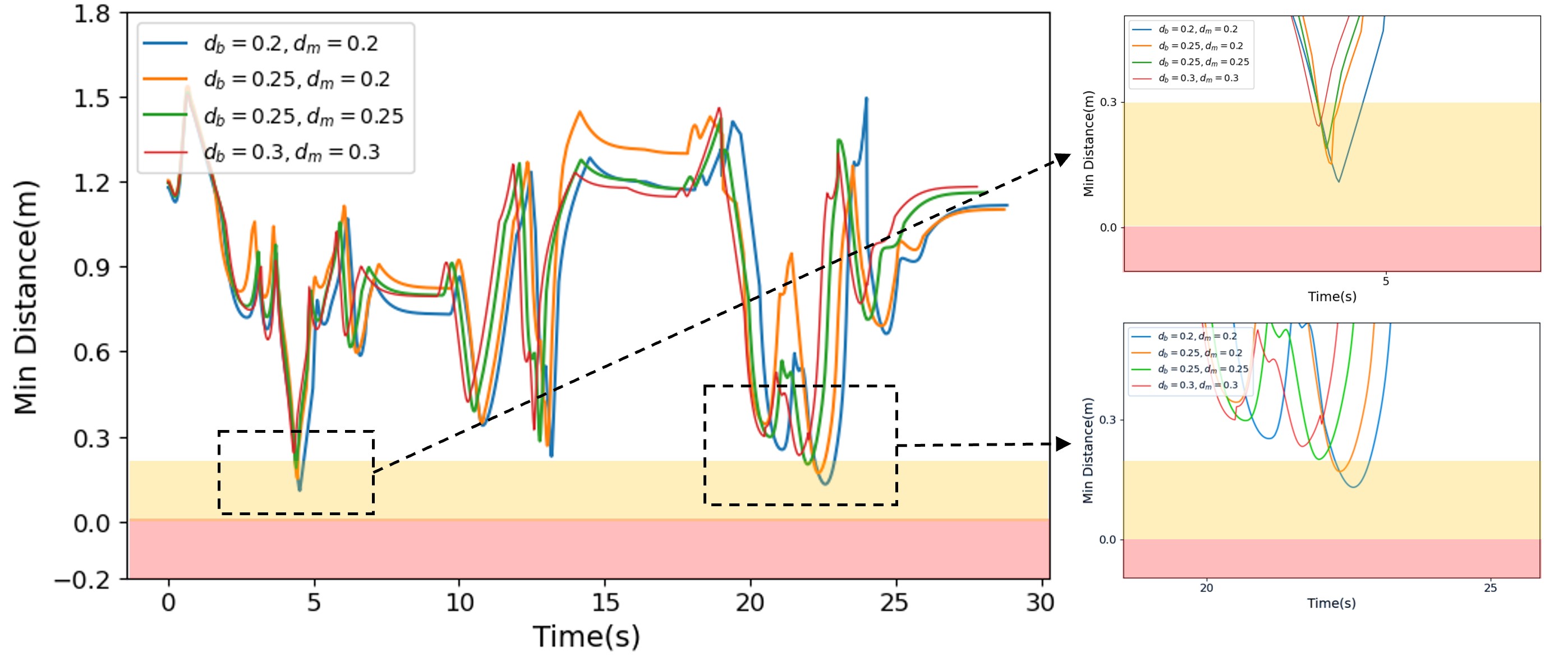}
	\caption{The minimum distance curve between the obstacles and the robot in Experiment 1, obtained using different values for the parameters $d_b$ and $d_m$.}
	\label{exp1}
\end{figure}

The experimental results are summarized in Table \ref{table_compare2}. In this study, we conducted 50 trials, where the robot’s initial $x$ and $y$ positions were randomly varied within the range $[-0.2, 0.2]$, and the orientation angle was varied within $[0, 2\pi)$. Each method is evaluated based on four metrics: computation frequency, total execution time, trajectory length, and minimum distance to obstacles. A minimum distance of less than zero indicates a collision (the system continues moving even after a collision). While the methods in \cite{Reis2021} and \cite{cir} prevent the system from the pseudo-equilibrium problem, they are not effective in handling dynamic obstacles. Moreover, since the method in \cite{cir} always selects a fixed direction to bypass obstacles, the resulting  path tends to be unnecessarily cumbersome. In \cite{mpc_mm}, the limited calculation frequency of the MPC prevents effective response to fast-moving dynamic obstacles. Moreover, using only a single CBF constraint leads to fall in the pseudo-equilibrium points and not reach the target. Add the ACI of mobile base solve this problem, but the manipulator still have collisions with dynamic obstacles. In contrast, the  safety constraints of the combined CBF and ACI (b, m) are demonstrated to effectively avoid dynamic obstacles, maintaining a safe distance. Furthermore, Fig. \ref{exp1} shows that higher $d_b$ or $d_m$ triggers the ACI earlier, prompting the obstacle avoidance action sooner and enabling the robot to avoid obstacles with a safer margin. These parameter settings can be adjusted according to specific safety margin requirements. Notably, our controller's parameters remain effective when adjusted within a certain range. 

\begin{table}[t]
    \centering
    \caption{Simulation results in 50 times}
    \resizebox{\linewidth}{!}{  
    \renewcommand{\arraystretch}{1.3} 
    \begin{tabular}{ccccc}
        \hline
        Methods & Frequency & Time & Length  & Min Obs Dis \\
        \hline
        Double CBF \cite{Reis2021}  & 131.14 Hz & 33.42 s & 30.16 m  & -0.09 m\\
        Circulation-CBF \cite{cir} & 136.73 Hz  & 37.11 s  & 33.59 m  & -0.02 m \\
        MPC \cite{mpc_mm} & 9.58 Hz &  27.52 s & \textbf{27.62} m & -0.08 m \\
        CBF				& \textbf{197.53 Hz} &  -  &  -  &  - \\
        CBF+ACI (b)      & {172.67 Hz} & 30.24 s  & 29.96 m & -0.03 m \\ 
        CBF+ACI (b,m)    & 105.26 Hz  &  \textbf{27.35 s}  & 27.81 m	& \textbf{0.05 m} \\
        \hline
    \end{tabular}
    }
    \label{table_compare2}
\end{table}

\subsection{Experiment 2: Real-World Scene with Flying Balls}
In this section, we test if the robot can autonomously avoid obstacles and reach the target point in the real world. We construct a large-scale scene featuring static obstacles, railings, and more. The robot will begin from its initial position at the EE $(-1.5,-1.5,1.1)$ and navigate towards the target point $(2.6, 0.4, 0.8)$. And we would throw some balls where Kalman filters are used for the ball's state estimation. Given the ball's potential for unexpected situations like collisions and bounces with static obstacles, we will promptly filter it directly once it drops below a certain height. Especially, we use the parameter $d_m=0.5$ when avoiding the fly ball and other parameters remain unchanged.

\begin{figure*}[t]
	\centering
	\includegraphics[width=\textwidth]{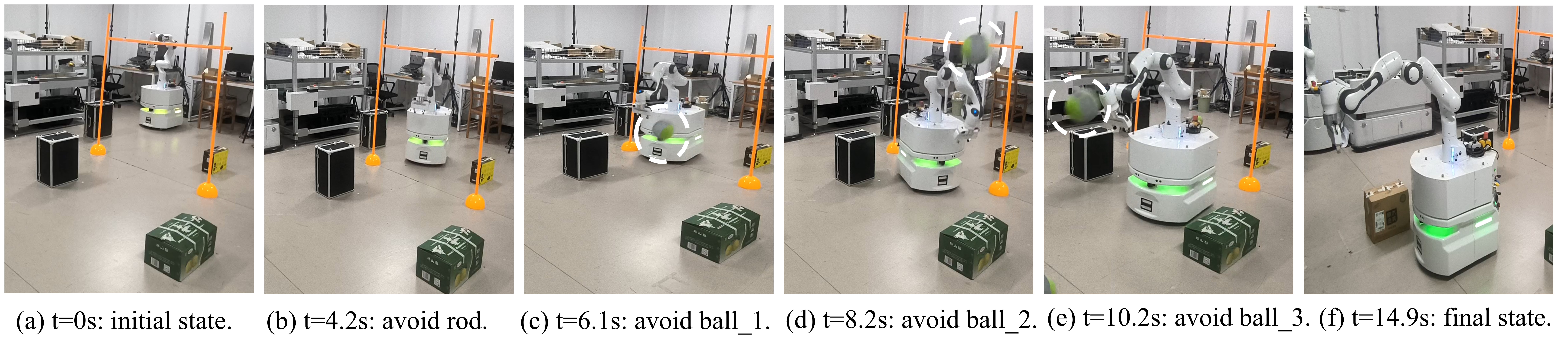}
	\caption{The physical robot is located in a large scene to avoid obstacles on the ground and dodge incoming balls in Experiment 3.}
	\label{exp3_show}
\end{figure*}

\begin{figure*}[t]
	\centering
	\includegraphics[width=\textwidth,,height=3cm]{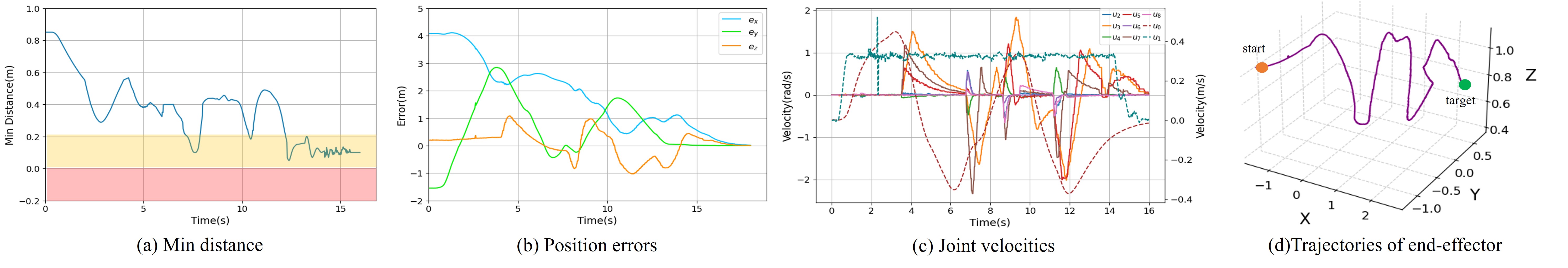}
	\caption{Min distance, position errors, joint velocities, and trajectories of end-effector of our controller in Experiment 2. }
	\label{exp3}
\end{figure*}

Fig.\ref{exp3_show} shows the the physical robot of experiment 2. Our robot successfully avoids ground obstacles and unexpected balls thrown from a distance while smoothly reaching the target point. During this process, balls traveling at speeds of up to 2m/s were thrown to the robot. As illustrated in Fig. \ref{exp3}, the minimum distance between the robot and obstacles is always greater than zero. The end effector's position errors eventually converge to zero, and the joint velocities remain within the constraints. Moreover, the fact that both the base and joint velocities reach zero simultaneously indicates that the MM no longer reaches the target through decoupled control, but rather through a graceful, whole-body coordinated motion.

\begin{figure*}[t]
	\centering
	\includegraphics[width=\textwidth,height=3cm]{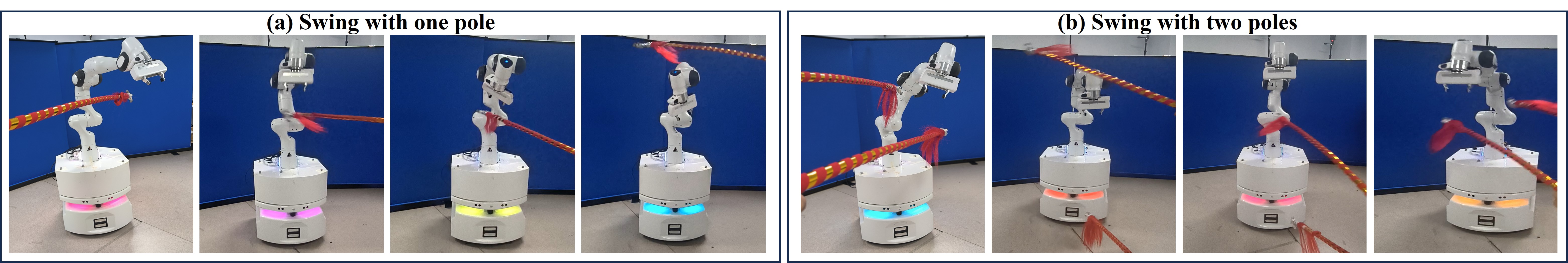}
	\caption{The robot motions using our controller in Experiment 3a)/3b). With one or two poles approaching the robot in different movements and directions, the robot performs a quick dodge to avoid collision.}
	\label{exp2_show}
\end{figure*}

\begin{figure*}[h]
	\centering
	\begin{subfigure}[b]{0.48\textwidth}
		\centering
		\includegraphics[width=\textwidth,height=2.5cm]{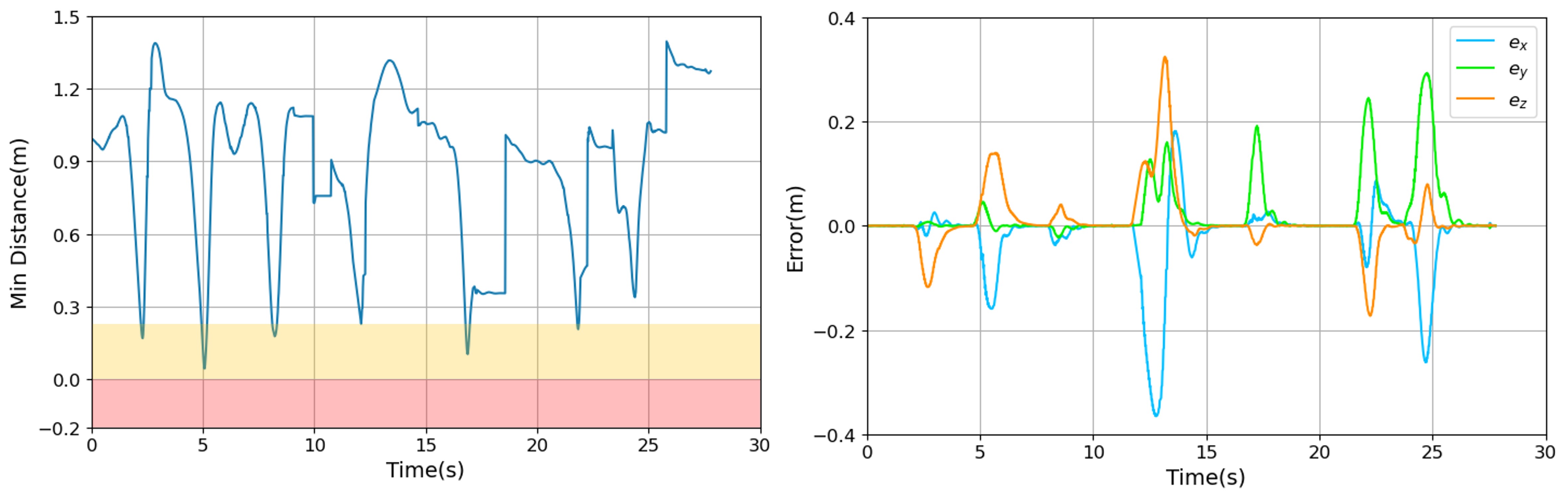}
		\caption{Experiment 3a)}
		\label{exp2_a}
	\end{subfigure}
	\hspace{0.01\textwidth} 
	\begin{subfigure}[b]{0.48\textwidth}
		\centering
		\includegraphics[width=\textwidth,height=2.5cm]{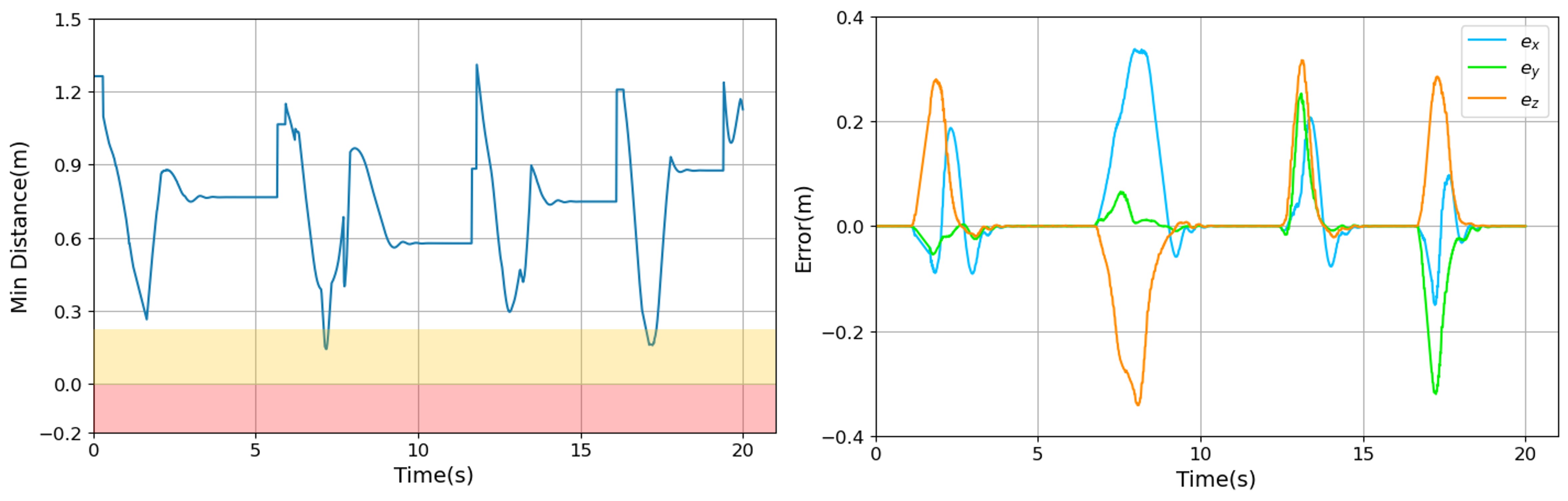}
		\caption{Experiment 3b)}
		\label{exp2_b}
	\end{subfigure}
	\caption{Min distance between MM and obstacles, and the position error between the EE and the given target point in Experiment 3a)/3b). }
	\label{exp2}
\end{figure*}

\subsection{Experiment 3: Avoidance Agility Test for Reactive Controller}
For reactive controllers, the ability to quickly recover during continuous dynamic obstacle avoidance serves as a strong validation of their responsiveness and agility. 
To simulate successive fast-moving obstacles, we employ one or two hand-held poles held by human as dynamic obstacles. A reflective marker is placed at the tip of the pole, allowing it to be passively tracked by the motion capture system. Once tracked, we model it as a virtual sphere with a radius of 4 cm centered at the marker's position. The controller then computes the collision function $h$ as usual. The proposed controller's performance will be compared with \cite{neo}, which is a also a reactive controller for collision avoidance.  

\textit{a) Single pole}: In this scenario, one pole held by human moves towards the robot at approximately 1.5 m/s. The robot must utilize its full range of joint motions to avoid the obstacle swiftly. Once safety is ensured, the EE of the robot is required to return to the target point as quickly as possible.

\textit{b) Two poles}: This scenario extends the complexity by introducing two poles  approaching the robot simultaneously from different directions. 

\cite{neo} performed poorly in Experiment 3a), primarily because its reactivity relies on directly eliminating the relative velocity with respect to the obstacle, which often leads to infeasibility. For our controller in experiment 2a)2b), we display the robot motion in Fig. \ref{exp2_show}. We achieved an effect that the above controller could not accomplish. We attempted to approach the robot from different directions—front, oblique angles, and different movements-stabbing, swinging, to provoke collisions. We also conducted experiments where both the base and the manipulator encountered obstacles simultaneously in Experiment 3b). Throughout this process, our controller effectively navigated the robot to avoid obstacles, promptly returning it to the target point once safety was ensured. The entire sequence was autonomously managed by the robot through our controller.  Fig. \ref{exp2} (a) and Fig. \ref{exp2} (b) represent the avoidance of a single pole and two poles, respectively. It is evident that the minimum distance between the robot and the obstacle always remains above zero, and any disturbances on the EE's position caused by the obstacle are swiftly recovered.

\section{Conclusions}
In this letter, we present a SEWB control that ensures both internal and external collision-free motion for mobile manipulators. We propose an novel ACI approach, which combines CBF to establish safety assurance constraints with a primary optimization QP. And we proved that ACI and CBF wound not conflict with each other. Our experiments demonstrate that it performs more efficiently and robustly than existing methods. SEWB solves the classical pseudo-equilibrium point problems generated by conventional CBF-based methods and enables the capacity that avoid dynamic obstacles with high agility.  However, it is worth noting that in the presence of significant system state disturbances or inaccurate environmental information, the safety constraints may become unreliable, especially for $\hat{\boldsymbol{\xi}}_{ob}$. In future work, we plan to incorporate robust estimation algorithms to address these challenges.

\bibliographystyle{IEEEtran}
\bibliography{ref}

\end{document}